\documentclass[10pt,twocolumn,letterpaper]{article}

\usepackage{iccv}              
\usepackage{graphicx}
\usepackage{amsmath}
\usepackage{amssymb}
\usepackage{booktabs}
\usepackage{adjustbox}
\usepackage{makecell}
\usepackage{multirow}
\usepackage{bbding}
\usepackage{color,soul}

\usepackage{xcolor}
\definecolor{iccvblue}{rgb}{0.21,0.49,0.74}
\usepackage[pagebackref,breaklinks,colorlinks,allcolors=iccvblue]{hyperref}

\def\paperID{15710} 
\def\confName{ICCV}
\def\confYear{2025}

\title{Causal Disentanglement and Cross-Modal Alignment for Enhanced Few-Shot Learning}

\author{Tianjiao Jiang, Zhen Zhang, Yuhang Liu\thanks{Corresponding author. Email: yuhang.liu01@adelaide.edu.au} , Javen Qinfeng Shi\\
Australian Institute for Machine Learning, The University of Adelaide, Australia\\
{\tt\small \{tianjiao.jiang, zhen.zhang02, yuhang.liu01, javen.shi\}@adelaide.edu.au}
}

\begin{document}
\maketitle
\begin{abstract}
Few-shot learning (FSL) often requires effective adaptation of models using limited labeled data.
However, most existing FSL methods rely on entangled representations, requiring the model to implicitly recover the unmixing process to obtain disentangled representations using only limited supervision, which hinders effective adaptation.
Recent theoretical studies show that multimodal contrastive learning methods, such as CLIP, can disentangle latent representations up to linear transformations.
In light of this, we propose the Causal CLIP Adapter (CCA), a novel framework that explicitly disentangles visual features extracted from CLIP using unsupervised Independent Component Analysis (ICA).
This removes the need to learn the unmixing process from the labeled data, thereby reducing the number of trainable parameters and mitigating overfitting. Taking a step further, while ICA can obtain visual disentangled representations, it may also disrupt CLIP’s intra- and inter-modal alignment. To counteract this, CCA further leverages CLIP’s inherent cross-modal alignment by enhancing it in two ways: unidirectionally, through fine-tuning a CLIP-based text classifier, and bidirectionally, via a cross-attention mechanism that enriches visual and textual representations through mutual interaction. Both uni-modal and cross-modal classification outputs can be effectively combined linearly to improve classification accuracy. Extensive experiments on 11 benchmark datasets demonstrate that our method consistently outperforms state-of-the-art approaches in terms of few-shot performance and robustness to distributional shifts, while maintaining computational efficiency. Code will be available at \url{https://github.com/tianjiao-j/CCA}. 
\end{abstract}


\section{Introduction}
\label{sec:intro}

\begin{figure}[ht]
    \centering
        \includegraphics[width=0.475\textwidth]{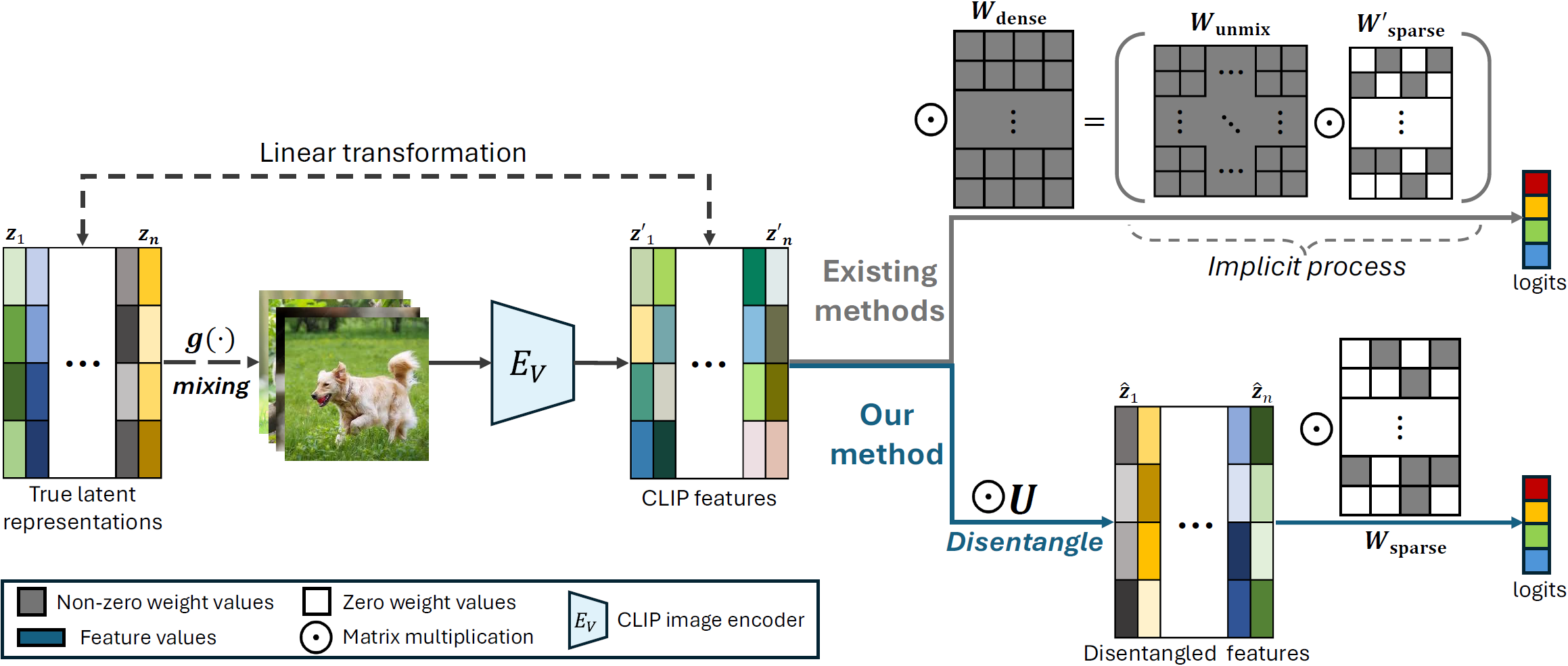}
    \caption{In the causal generative model, we assume that images are generated by an unknown mixing function $\mathbf{g}(\cdot)$ from a set of latent variables ${\mathbf{z}_1, \mathbf{z}_2, \ldots, \mathbf{z}_n}$. The outputs of the CLIP image encoder are entangled representations where each $\mathbf{z}'_i$ is a linear mixture of the true latent variables. Existing methods directly use these entangled CLIP features for downstream tasks, which necessitates implicit learning of a dense weight matrix $\mathbf{W}_{\text{dense}}$ that can be regarded as the product of an unmixing matrix $\mathbf{W}_{\text{unmix}}$ and a sparse matrix $\mathbf{W}^{\prime}_{\text{sparse}}$. In contrast, our method first disentangles the CLIP features using an unmixing matrix $\mathbf{U}$ obtained by ICA, resulting in disentangled features where $\mathbf{\hat{z}}$ approximates $\mathbf{z}$ up to permutation and scaling. These disentangled features are then used for downstream tasks, requiring only the learning of a sparse weight matrix $\mathbf{W}_{\text{sparse}}$.}
    \label{fig:disentangle}
\end{figure}

Conventional machine learning models typically require large amounts of labeled data to train parameter-rich architectures effectively and achieve strong performance. However, in many real-world applications, labeled data can be scarce, costly, or impractical to obtain. Few-shot learning (FSL) addresses this limitation by adapting models to new tasks with only a small number of samples, mimicking the human ability to learn from limited data and generalize to unseen instances \cite{wang2020generalizing,sung2018learning,snell2017prototypical}. 
Two primary approaches are commonly used in FSL: meta-learning \cite{chen2021meta,vinyals2016matching,andrychowicz2016learning} and transfer learning \cite{yu2020transmatch,shen2021partial,hussain2019study}. In meta-learning, a model is trained using a support set to develop the ability to learn from only a few samples, and its performance is evaluated on a query set. In contrast, transfer learning leverages pre-trained models trained on large datasets to capture rich semantic representations and rapidly adapt to new tasks. These pre-trained models range from smaller architectures, such as ResNet-50 \cite{he2016deep}, to larger multi-modal frameworks such as CLIP \cite{radford2021learning}. 
Despite their effectiveness, both FSL approaches often fail to generalize \cite{snell2017prototypical,sun2019meta}, primarily due to the challenge of implicitly learning the unmixing of original representations to obtain disentangled features using only limited labeled data. See \cref{fig:disentangle} for details.

Recent advances in identifiability analysis have provided significant theoretical insights into causal disentanglement from various perspectives, such as data diversity \cite{hyvarinen2019nonlinear,khemakhem2020variational,liu2022identifying,liuidentifiable,liu2022identifying1,liu2024identifiable,liu2025predict} and contrastive learning \cite{radford2021learning,jia2021scaling,li2022blip,daunhawer2023identifiability,gresele2020incomplete,liu2024revealing,cai2024clap,cai2025value}. In particular, CLIP-like models trained via multi-modal contrastive learning have been shown to possess strong disentanglement potential, enabling the recovery of true latent variables up to linear or permutation transformations, depending on the underlying model assumptions \cite{liu2024revealing}. Building on this theoretical insight, rather than directly employing the representations learned by CLIP-like models in existing methods for FSL, we first extract disentangled representations from CLIP using Independent Component Analysis (ICA) to remove the linear transformation. These representations are then used for downstream FSL tasks. By isolating independent latent factors, the disentangled representations enable CLIP to more effectively learn the influence of individual factor on ground-truth labels, while requiring fewer parameters to be learned.

Taking a step further, while disentanglement via ICA enhances representation learning, it may also disrupt intra-modal alignment in CLIP features when applied to the image modality only. Moreover, fine-tuning only the cache model - a common practice in many existing methods \cite{gao2021clip,zhou2021coop,zhang2022tip} - may not be sufficient to achieve optimal cross-domain generalization ability. To address this, we build upon CLIP’s strong few-shot \cite{gao2021clip,zhou2021coop,zhang2022tip} and zero-shot learning capabilities \cite{khattak2023maple,martin2024transductive} by constructing a CLIP text classifier following the standard procedure \cite{radford2021learning}. This classifier is fine-tuned to unidirectionally enhance cross-modal alignment on downstream datasets. Beyond unidirectional fine-tuning, we further leverage CLIP’s inherent image-text alignment to improve feature representations bidirectionally. By employing a cross-attention mechanism which enables the model to capture richer semantic relationships between visual and textual modalities, we construct a refined text classifier from visual features and refine image features using textual features. These hybrid representations contribute additional logits that complement the original CLIP features. The final prediction is obtained through a linear combination of logits from the cache model, the CLIP text classifier, and the hybrid cross-modal features, allowing the model to effectively leverage both intra-modal and inter-modal alignment for more robust and accurate predictions. Experimental results demonstrate that our method consistently outperforms several state-of-the-art (SOTA) methods, highlighting the benefits of causal disentanglement and improved cross-modal alignment. 
Our contributions involve the following:
\begin{itemize}
    \item We introduce the Causal CLIP Adapter (CCA), which leverages CLIP and ICA to obtain disentangled latent representations for few-shot learning, enhancing adaptability to downstream distribution shifts.
    \item We propose addressing the limitations of uni-modal classification by enhancing the cross-modal alignment of CLIP features both unidirectionally and bidirectionally through fine-tuning and cross-attention.
    \item We assessed the effectiveness of our method on 11 widely used image classification datasets and conducted ablation studies to further analyze and characterize its performance.
    \end{itemize}

\begin{figure*}[ht]
  \centering
   \includegraphics[width=0.8\textwidth]{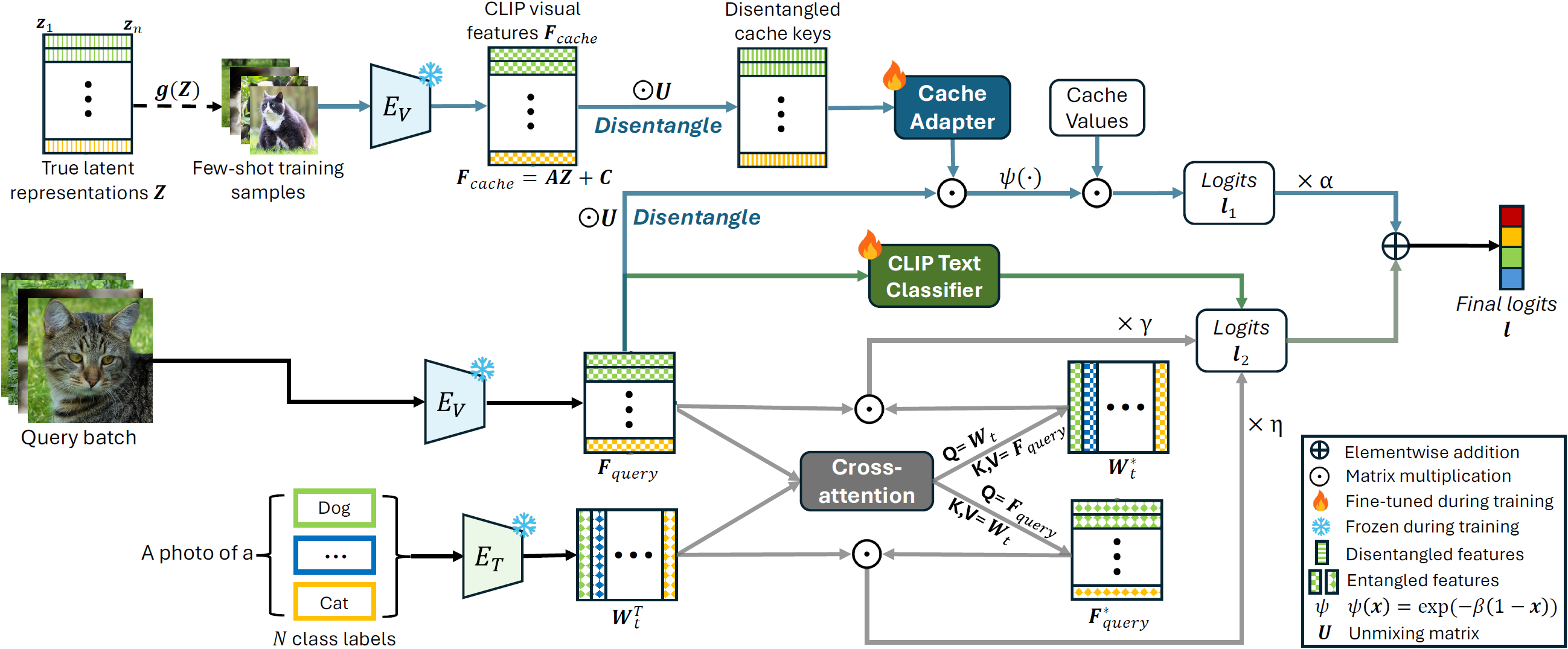}
   \caption{\textbf{Architecture of the Causal CLIP Adapter.} Assuming the few-shot training samples are generated from a set of latent vectors $\mathbf{Z}=\{\mathbf{z}_1,\mathbf{z}_2,...,\mathbf{z}_n\}$ by a non-linear function $\mathbf{g}(\cdot)$, the features $\mathbf{F}_{\text{cache}}$ outputted by CLIP visual encoder $E_V$ are related to the true generating latents by $\mathbf{F}_{\text{cache}}=\mathbf{AZ+C}$, where $\mathbf{A}$ is an orthogonal matrix and $\mathbf{C}$ is a constant offset. These entangled features are then disentangled using an unmixing matrix $\mathbf{U}$ obtained via FastICA, recovering the latent variables up to permutation and scaling. The resulting disentangled features serve as cache keys. During training, a cache adapter (initialized as the identity matrix) is fine-tuned to enhance alignment within the image modality. Its output is multiplied with fixed cache values - one-hot vectors representing the few-shot ground-truth labels - to produce logits $\mathbf{l}_1$, weighted by $\alpha$. In parallel, the CLIP text classifier, derived from the CLIP text encoder $E_T$, is fine-tuned to predict class labels by leveraging CLIP’s image-text alignment. Additionally, the image and text features are further fused using a cross-attention mechanism, resulting in two sets of hybrid features $\mathbf{W}_t^*$ and $\mathbf{F}_{\text{query}}^*$ for two more sets of predictions, weighted by $\gamma$ and $\eta$ respectively. The cross-modal alignment module produces logits $\mathbf{l}_2$. The logits from different components are combined linearly to generate the final logits.}
   \label{fig:model}
\end{figure*}


\section{Related Work}
\label{sec:related}

In recent years, large-scale multi-modal pre-training, which leverages vast amounts of multi-modal data to train models that learn from diverse information across modalities and align data between them, has gained popularity due to its strong generalization performance \cite{radford2021learning,jia2021scaling,li2022blip}. For example, CLIP \cite{radford2021learning} uses large-scale image-text pairs gathered from the internet to train a visual encoder and a textual encoder simultaneously. ALIGN \cite{jia2021scaling} employs a noisy dataset of image-text pairs to train a visual encoder and a textual encoder concurrently, though its encoder architecture differs from that of CLIP. These vision-language models effectively learn visual concepts from natural language supervision and project visual and textual features into a shared embedding space.
The pre-trained CLIP model has shown a remarkable domain generalization ability on data that it has never encountered before. This capability is highly beneficial for transfer learning, including zero-shot and few-shot learning \cite{radford2021learning}. CLIP-based few-shot learning approaches fall into three main categories: prompt-tuning, adapter-based methods, and methods based on additional knowledge.

Prompt-tuning methods for few-shot learning incorporate tunable parameters into the input of CLIP’s textual and/or visual encoders to better align images and text for downstream tasks. These methods include textual prompt-tuning, visual prompt-tuning, and multi-modal prompt-tuning.
For textual prompt-tuning, CoOp \cite{zhou2022learning} replaces hand-crafted prompts with trainable vectors, while CoCoOp \cite{zhou2022conditional} enhances generalization by conditioning prompts on images. ProGrad \cite{zhu2023prompt} refines gradient updates to preserve CLIP's prior knowledge and reduce overfitting. 
In visual prompt-tuning, trainable parameters can be incorporated into the transformer encoder \cite{jia2022visual} or implemented as image paddings \cite{bahng2022exploring}. Multi-modal prompt-tuning jointly learns textual and visual prompts to enhance cross-modal alignment. For example, MaPLe (Multi-modal Prompt Learning) \cite{khattak2023maple} conditions visual prompts on textual ones for better synergy, UPT (Unified Prompt Tuning) \cite{zang2022unified} decomposes a shared prompt into separate textual and visual components.

Besides prompt-tuning, adapter-based methods have been widely explored to improve the generalization of multi-modal foundation models. CLIP-Adapter \cite{gao2024clip} introduces a linear adapter, freezing the CLIP backbone and combining few-shot knowledge with CLIP’s zero-shot predictions via residual connections. Tip-Adapter \cite{zhang2022tip} constructs a cache model using few-shot training data, which makes predictions based on image-image similarities and is integrated with CLIP’s text classifier via a scaling factor. APE (Adaptive Prior rEfinement) \cite{zhu2023not} improves intra-class similarity and inter-class variance by pruning CLIP feature channels and refining representations via a residual matrix. FD-Align (Feature Discrimination Alignment) \cite{song2024fd} preserves the consistency of spurious features during fine-tuning with a loss term, improving robustness to spurious correlations and boosting both in-distribution and out-of-distribution performance.
To balance prior knowledge retention and flexibility, TaskRes (Task Residual Tuning) \cite{yu2023task} introduces a prior-independent task residual, incorporating tunable parameters for downstream tasks while keeping the prior module fixed.

Few-shot learning methods that incorporate prior knowledge have gained popularity for their superior performance. This knowledge can come from pre-trained models or test set information.
For CLIP combined with pre-trained models, the additional models could complement CLIP or address the limitations of few-shot learning. CaFo (Cascade of Foundation models) \cite{zhang2023prompt}, SuS-X \cite{udandarao2023sus}, and DeIL (Direct-and-Inverse CLIP) \cite{shao2024deil} use generative models such as DALL-E \cite{ramesh2021zero} and Stable Diffusion \cite{rombach2022high} to generate additional training samples. CaFo and AMU-Tuning \cite{tang2024amu} incorporate self-supervised models such as DINO \cite{caron2021emerging} and MoCo v3 \cite{chen2021empirical} to generate image features.
Another approach to improve few-shot learning performance involves leveraging relationships between test samples (i.e., transductive few-shot learning). DMN (Dual Memory Networks) \cite{zhang2024dual} assigns pseudo-labels and constructs memory networks with high-confidence samples. Transductive-CLIP \cite{martin2024transductive} refines predictions via Dirichlet-based modeling and iterative self-calibration.
However, pre-trained models such as DINO come with high upfront training costs, and real-world test samples may not arrive in batches, making these approaches impractical for our study.

Although the above methods have improved CLIP's zero-shot and few-shot performance, the features used for prediction remain entangled, complicating optimization and limiting further performance gains.


\section{Proposed Method}
\label{sec:method}
We introduce the Causal CLIP Adapter (CCA), designed to enhance few-shot learning (FSL) performance through both intra-modal and inter-modal feature alignment. Section~\ref{disentanglement} explains the role of disentanglement in FSL and how our model leverages disentangled representations. Section~\ref{alignment} details our approach to improve cross-modal alignment, further enhancing model performance. Section~\ref{prediction} outlines the final prediction process and describes the training methods used for our model.

\subsection{Disentanglement for Few-shot Learning}
\label{disentanglement}
In the causal generative process, we assume that a set of independent latent variables $\mathbf{z}$ governs the fundamental properties of the observed data, such as the semantics and style of an image. When entangled representations are used to train a classifier, each latent variable incorporates a mixture of factors, such as specific content and style properties, resulting in dense weights with most values being non-zero. In fact, the classifier will have to implicitly learn the unmixing procedure. In FSL, this entanglement increases the risk of overfitting, as the limited training data forbids the accurate estimation of the unmixing procedure due to the large number of parameters. Conversely, with disentangled representations, the classifier only needs to learn a few parameters as it is often observed that only a few latent factors may have an effect on the labels, while others, such as those encoding style information, may be irrelevant to the label, and the corresponding weights for these variables would naturally be zero. Thus, in FSL, where training data is scarce, accurately capturing the true latent variables $\mathbf{z}$ becomes particularly critical for reducing the number of parameters needed to fine-tune a classifier and mitigating overfitting.

\subsubsection{Disentanglement by FastICA}
For $M$ dimensional CLIP features living on a $M-1$ dimensional unit hypersphere, the maximum number of independent components is $M-1$ since hyperspheres lack one degree of freedom. Therefore, CLIP features need to be disentangled to obtain the independent latent variables. As demonstrated in \cite{liu2024revealing}, CLIP features represent a linear mixture of the true latent variables. Specifically, let $\mathbf{z}$ denote the true latent variables for an image in the shared latent space, and $\mathbf{z}'$ represent the latent variables obtained by feeding the image into the CLIP visual encoder. For hyperspheres, their relationship is given by $\mathbf{z}' = \mathbf{Az} + \mathbf{c}$, where $\mathbf{A}$ is an orthogonal matrix, and $\mathbf{c}$ is a constant vector.

In essence, $\mathbf{z}$ can be identified up to a trivial linear transformation in hyperspherical spaces. Once $\mathbf{z}'$ is obtained, the true latent variables $\mathbf{z}$ can be recovered using linear independent component analysis (ICA), assuming that at most one of the true latent variables in $\mathbf{z}$ follows a Gaussian distribution. For images, the true latent variables are often highly non-Gaussian, making ICA algorithms such as FastICA \cite{hyvarinen1999fast}, which maximize non-Gaussianity, particularly effective for recovering them. Additionally, FastICA is well-suited for handling the high dimensionality of CLIP features, offering fast convergence and ensuring scalability for large datasets.

\subsubsection{Disentangled Cache Model}
For a $K$-shot $N$-class FSL problem for image classification, where each of the $N$ categories contains $K$ labeled images, we will use $\mathbf{I}_{\text{train}}$ and $\mathbf{L}_{\text{train}}$ to denote the whole collection of $N\times K$ training images and labels. Then following \cite{zhang2022tip} the training set is used to construct cache keys and labels as follows, 
\begin{align}
\label{cache}
    &\mathbf{F}_{\text{cache}} = \mathrm{Visual Encoder}(\mathbf{I}_{\text{train}}), \notag \\
    &\mathbf{L}_{\text{cache}} = \mathrm{One Hot}(\mathbf{L}_{\text{train}}), 
\end{align}
where $\mathbf{F}_{\text{cache}} \in \mathbb{R}^{NK \times C}$, $\mathbf{L}_{\text{cache}} \in \mathbb{R}^{N \times NK}$. Then for a set of $B$ query images $\mathbf{I}_{\text{query}}$, their features are obtained as follows
\begin{align}
&\mathbf{F}_\text{query} = \mathrm{Visual Encoder}(\mathbf{I}_{\text{query}}),
\end{align}
where $\mathbf{F}_\text{query} \in \mathbb{R}^{B \times C}$. 
Next, we leverage CLIP's disentanglement ability and ICA to transform the cache model into a disentangled cache model. Specifically, the cache keys are disentangled using an unmixing matrix $\mathbf{U} \in \mathbb{R}^{C \times M}$, derived through FastICA\footnote{Due to the limited number of training samples for FSL, we extract the unmixing matrix $\mathbf{U}$ from a subset of ImageNet \cite{deng2009imagenet} and use it for all downstream datasets.} \cite{hyvarinen1999fast} to transform the $C$-dimensional CLIP features into $M$-dimensional disentangled representations, where $C\geq M$. The disentangled cache keys are then passed through a trainable cache adapter with weights $\mathbf{W}_c \in \mathbb{R}^{M \times M}$ initialized as identity. This transformation is represented in \cref{ica}, where $\mathbf{F}_{\text{cache}}^{d} \in \mathbb{R}^{NK \times M}$ denotes the disentangled cache keys. The features of the query images are also disentangled using the same unmixing matrix $\mathbf{U}$.
\begin{align}
\label{ica}
    &\mathbf{F}_{\text{cache}}^{d} = \mathbf{F}_{\text{cache}} \mathbf{U}\mathbf{W}_c, \notag \\
    &\mathbf{F}^{d}_\text{query} = \mathbf{F}_\text{query} \mathbf{U}.
\end{align}
Once the disentangled cache model is constructed, we can compute the scaled similarities between disentangled image features as in \cref{sim},
\begin{align}
\label{sim}
    \mathbf{S} 
       =  \exp{(-\beta(1 - \mathbf{F}^{d}_\text{query} {\mathbf{F}_\text{cache}^d}^\top))},
\end{align}
where $\mathbf{S} \in \mathbb{R}^{B \times NK}$ contains the similarity scores of each query image to each few-shot training sample in the disentangled feature space, and $\beta$ controls the smoothness of the similarities. The logits for intra-modal alignment $\mathbf{l}_{1}\in \mathbb{R}^{B\times N}$ are subsequently obtained by \cref{l1}.
\begin{align}
\label{l1}
    \mathbf{l}_1=\mathbf{S} \mathbf{L}^\top_{\text{cache}}.
\end{align}

\subsection{Cross-modal Alignment}
\label{alignment}
To align text representations with image representations, we fine-tune the CLIP text classifier initialized with CLIP text features $\mathbf{W}_t \in \mathbb{R}^{N \times C}$ which is obtained by passing each class label into the CLIP text encoder. This process ensures that the text features are more closely aligned with the corresponding image features in the shared embedding space. By performing this fine-tuning, we unidirectionally align text features to image features, thereby reducing the semantic gap between the two modalities. This alignment improves the contextual relevance of the text features and enables the classifier to better associate textual descriptions with their visual counterparts, leading to enhanced prediction accuracy.

To further bridge the gap between image and text representations, we leverage CLIP's inherent image-text alignment and propose fusing image and text features through cross-attention mechanisms. Cross-attention enables the model to capture rich contextual dependencies between the two modalities, allowing for a bidirectional exchange of information. This results in a text classifier enriched with image features and image features augmented by text features. The hybrid features strengthen the connection between textual and visual modalities, ultimately improving the model's performance on downstream tasks.

Specifically, given the original CLIP visual features of the current batch $\mathbf{F}_\text{query}$ as queries ($\mathbf{Q}$) and the CLIP text features $\mathbf{W}_t$ as keys ($\mathbf{K}$) and values ($\mathbf{V}$). We obtain a text classifier enriched with image features by \cref{W_t_star}.
\begin{align}
\label{W_t_star}
    &\mathbf{W}_t^*=(\text{softmax}(\mathbf{W}_t\mathbf{F}_{\text{query}}^\top)\mathbf{F}_{\text{query}})^\top,
\end{align}
where $\mathbf{W}_t^* \in \mathbb{R}^{C \times N}$.

Similarly, we obtain the image features augmented by text features using $\mathbf{W}_t$ as queries ($\mathbf{Q}$) and $\mathbf{F}_{\text{query}}$ as keys ($\mathbf{K}$) and values ($\mathbf{V}$)\footnote{During testing, we use $\mathbf{F}_{\text{cache}}$ as $\mathbf{K}$ and $\mathbf{V}$ for this step.} following \cref{f_train_star}.
\begin{align}
\label{f_train_star}
    &\mathbf{F}_{\text{query}}^*=\text{softmax}(\mathbf{F}_{\text{query}} \mathbf{W}_t^\top)\mathbf{W}_t
\end{align}
where $\mathbf{F}_{\text{query}}^* \in \mathbb{R}^{B \times C}$. 

The logits for cross-modal alignment $\mathbf{l}_2$ is given by \cref{l2}.
\begin{align}
\label{l2}
    &\mathbf{l}_2=\mathbf{F}_\text{query} \mathbf{W}_t^\top + \gamma \mathbf{F}_\text{query} \mathbf{W}_t^* + \eta \mathbf{F}_{\text{query}}^* \mathbf{W}_t^\top,
\end{align}
where $\gamma$ and $\eta$ are balancing factors.

\subsection{Final Prediction and Fine-tuning}
\label{prediction}
The final logits $\mathbf{l}$ for the query images $\mathrm{I}_{\text{query}}$ are computed as follows.
\begin{align}
\label{logits}
    &\mathbf{l}=\alpha \mathbf{l}_1 +\mathbf{l}_2.
\end{align}

The proposed CCA can be fine-tuned on the few-shot training set while freezing the CLIP encoders, which is termed CCA-FT. It uses cross-entropy loss for image classification, and the model is updated via stochastic gradient descent (SGD). Specifically, we initialize the cache adapter weights $\mathbf{W}_c$ with an identity matrix and apply $\ell_1$ regularization during training to induce sparsity in $\mathbf{W}_c$. In the meantime, we fine-tune the CLIP text classifier initialized with $\mathbf{W}_t$ at a different learning rate. Fine-tuning the weights of the cache adapter enhances the intra-modal (i.e., image-image) alignment, which would result in more accurate similarity estimation. Updating the weights of the CLIP text classifier improves image-text alignment, thereby increasing the prediction accuracy. For most downstream datasets, CCA-FT only takes 20 epochs of training to achieve state-of-the-art (SOTA) performance, indicating a superior convergence speed.


\newcommand{\figAccs}{
\begin{figure*}[ht]
    \centering
        \includegraphics[width=\textwidth]{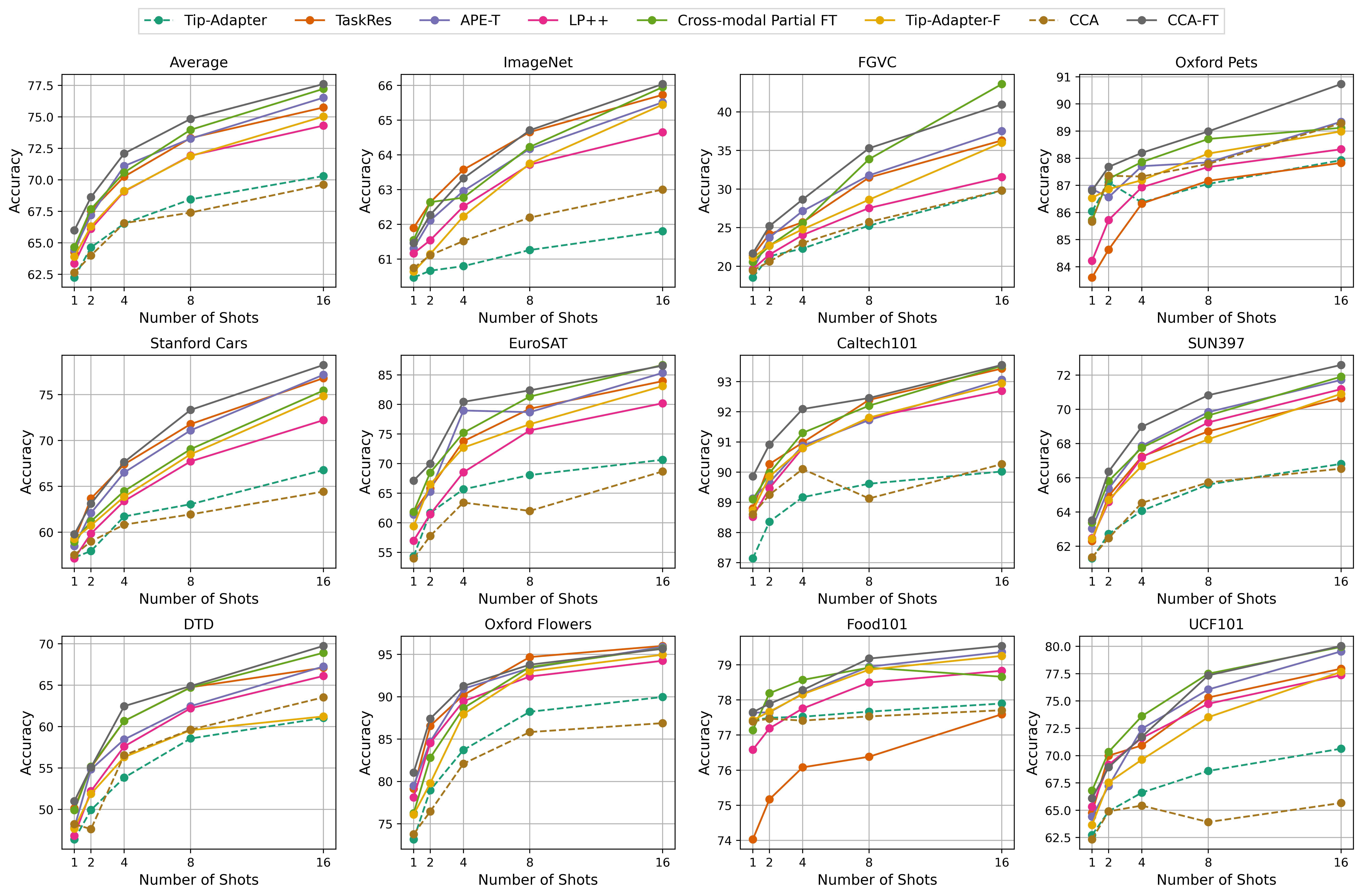}
    \caption{{\bf A comparison of top-1 classification accuracy between the CCA(-FT) and SOTA methods over 11 datasets, along with their average performance}. The x-axis indicates the number of training samples used per class from the downstream datasets. The y-axis indicates the classification accuracy (\%).}
    \label{fig:11_datasets}
\end{figure*}
}

\newcommand{\figHP}{
\begin{figure}[ht]
     \centering
     \begin{subfigure}[b]{0.35\textwidth}
         \centering
         \includegraphics[width=\textwidth]{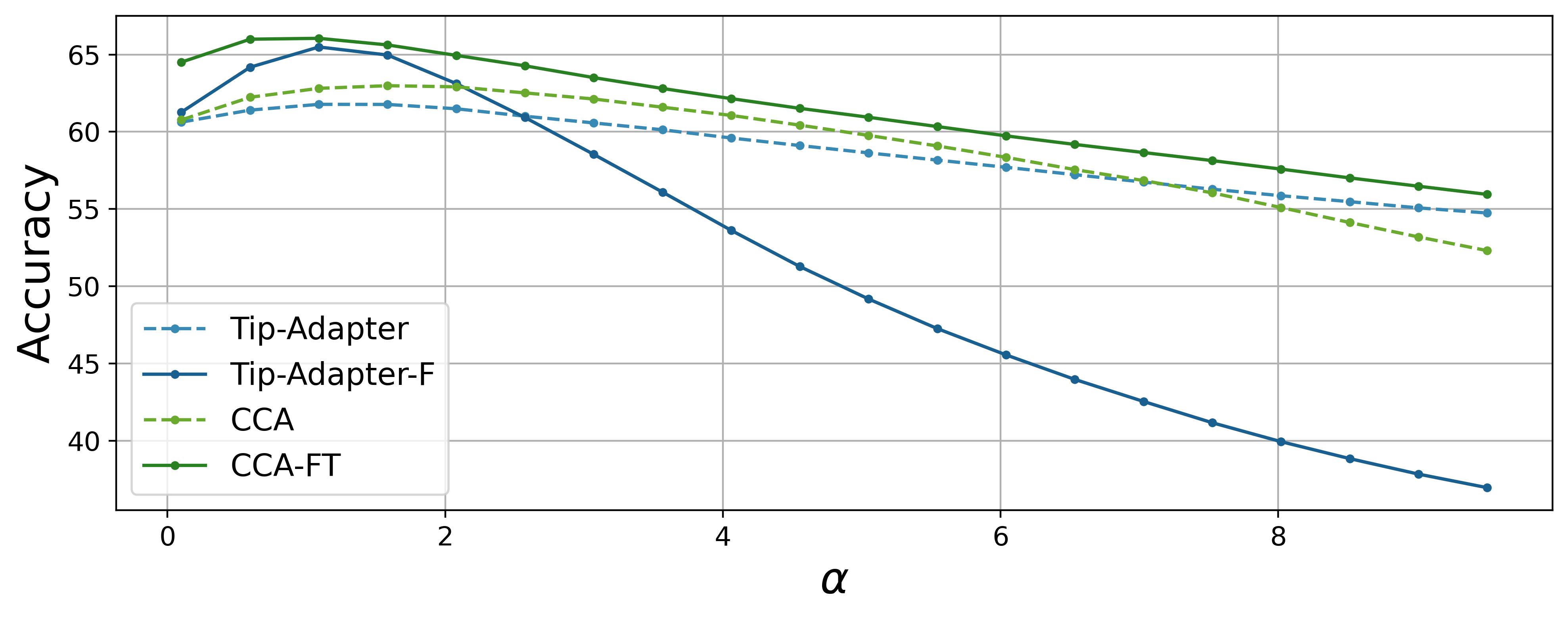}
         \caption{Sensitivity to $\alpha$.}
         \label{fig:alpha}
     \end{subfigure}
     \begin{subfigure}[b]{0.35\textwidth}
         \centering
         \includegraphics[width=\textwidth]{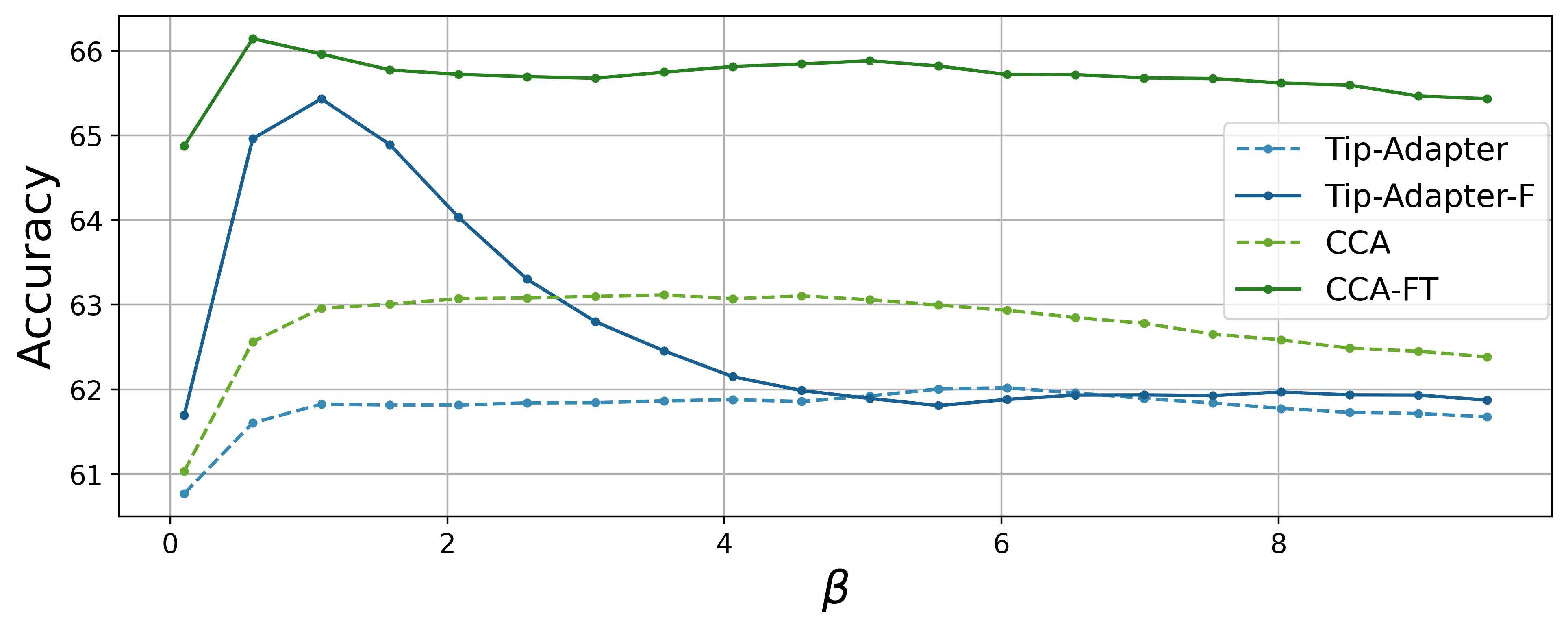}
         \caption{Sensitivity to $\beta$.}
         \label{fig:beta}
     \end{subfigure}
     \begin{subfigure}[b]{0.35\textwidth}
         \centering
         \includegraphics[width=\textwidth]{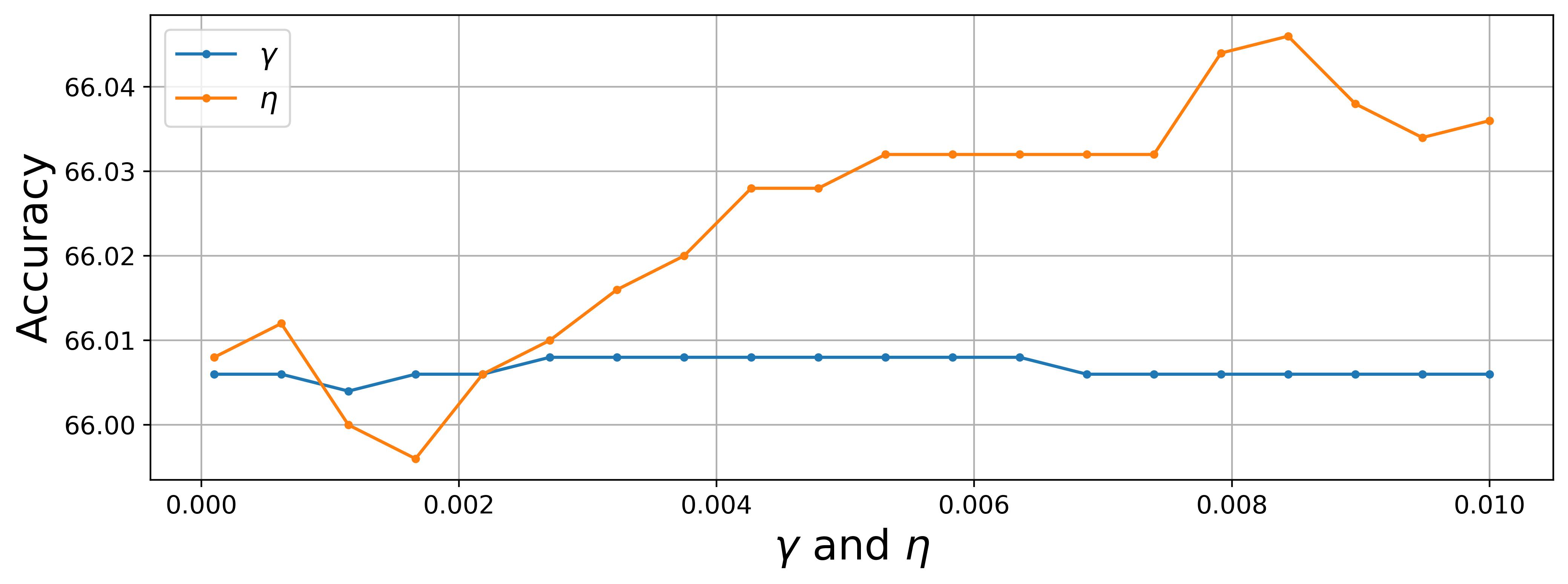}
         \caption{Sensitivity to $\gamma$ and $\eta$ for CCA-FT.}
         \label{fig:gamma_eta}
     \end{subfigure}
     \label{fig:hp}
     \caption{Sensitivities to the balancing factors $\alpha$, $\gamma$, $\eta$ and the smoothing factor $\beta$ for CCA(-FT) and Tip-Adapter(-F). The accuracy here represents 16-shot classification accuracy (\%) on ImageNet.}
\end{figure}
}

\newcommand\tabcaption{\def\@captype{table}\caption}

\newcommand{\tabTime}{
\begin{table}[t]
\small
\centering
\begin{adjustbox}{width=\linewidth}
	\begin{tabular}{lcccccc}
	\toprule
		Models & Epochs & Time &  Accuracy &Gain\\ \midrule
		Zero-shot CLIP &0 &0 &60.33 &0\\
		CoOp & 200 &7.5h & 62.95  & +2.62\\
		Tip-Adapter & 0 & 0 & 61.80\ & +1.47\\
	    Tip-Adapter-F & 20 & 4.8min & 65.45 & +5.12\\
        \midrule
        CCA & 0 & 0 & 63.00 & +2.67\\
	    CCA-FT & 20 & 4.9min & 66.04 & +5.71\\
	\bottomrule
	\end{tabular}
\end{adjustbox}
\caption{Computation efficiency of various methods for 16-shot learning on ImageNet. The last column shows the performance improvement compared to zero-shot CLIP.}
\label{time}
\end{table}
}

\newcommand{\tabOOD}{
\begin{table}
\centering
\small
\begin{tabular}{l@{\hskip 0.05in}c@{\hskip 0.05in}ccccc}
\toprule
\multirow{2}{*}{Datasets} & \textbf{Source} &\multicolumn{4}{c}{\textbf{Target Datasets}} \\
\cmidrule(lr){2-2} \cmidrule(lr){3-6} 
& ImageNet  & \quad-V2 & -Sketch \\
\midrule
Linear-probe CLIP & 56.13  & \quad45.61 & 19.13\\
CoOp & 62.95 & \quad55.11 & 32.74  \\
Tip-Adapter & 61.80 & \quad54.36 & 35.71  \\
Tip-Adapter-F & 65.45 & \quad56.85 & 35.00  \\
\midrule
CCA & 63.00  & \quad55.39 & \textbf{36.60}  \\
CCA-FT & \textbf{66.04}  & \quad\textbf{57.32} & 35.87  \\
\bottomrule
\end{tabular}
\caption{{\bf Robustness to Distributional Shifts}. All the methods were trained on 16-shot ImageNet training set. Bold numbers indicate the best performance for the corresponding dataset.}
\label{tab:ood}
\end{table}
}

\newcommand{\tabNoise}{
\begin{table}
\centering
\small
\begin{tabular}{l@{\hskip 0.05in}c@{\hskip 0.05in}ccccc}
\toprule
Method & 1-shot & \quad2-shot & 4-shot & 8-shot & 16-shot \\ 
\midrule
Tip-Adapter & 50.26 & \quad51.07 & 52.15 & 54.03 & 56.01 \\
Tip-Adapter-F & 52.19 & \quad54.41 & 56.23 & 58.62 & 60.04 \\
\midrule
CCA & 51.79 & \quad54.68 & 56.42 & 57.36 & 59.23 \\
CCA-FT & 54.22 & \quad56.30 & 58.03 & 61.39 & 62.48 \\
\bottomrule
\end{tabular}
\caption{{\bf Robustness to Gaussian noise}. The values represent the average classification accuracies (\%) across the 11 datasets.}
\label{tab:noise}
\end{table}
}

\newcommand{\tabAttack}{
\begin{table}%
\centering
\small
\begin{tabular}{l@{\hskip 0.05in}c@{\hskip 0.05in}ccccc}
\toprule
Method & 1-shot & \quad2-shot & 4-shot & 8-shot & 16-shot \\ 
\midrule
Tip-Adapter & 18.24 & \quad18.94 & 20.01 & 21.91 & 25.49 \\
Tip-Adapter-F & 19.34 & \quad21.50 & 24.23 & 28.50 & 33.21 \\
\midrule
CCA & 20.14 & \quad22.17 & 24.63 & 27.78 & 31.84 \\
CCA-FT & 21.44 & \quad24.17 & 26.50 & 30.53 & 35.48 \\
\bottomrule
\end{tabular}
\caption{{\bf Robustness to adversarial attack}. The values represent the average classification accuracies (\%) across the 11 datasets.}
\label{tab:attack}
\end{table}
}

\newcommand{\tabAblations}{
\begin{table}
\centering
\small
\begin{tabular}{l@{\hskip 0.09in}c@{\hskip 0.09in}c@{\hskip 0.09in}c@{\hskip 0.09in}c@{\hskip 0.09in}c}
\toprule
Ablation & 1-shot & 2-shot & 4-shot & 8-shot & 16-shot \\ 
\midrule
Complete model & 66.00 & 68.62 & 72.10 & 74.84 & 77.60 \\ 
\midrule
1) w/o ICA & 64.94 & 67.26 & 70.34 & 72.86 & 75.95 \\ 
2) fixed cache adapter & 65.16 & 67.75 & 70.78 & 73.12 & 75.75 \\ 
3) fixed text classifier & 65.10 & 67.96 & 71.32 & 74.04 & 77.00 \\ 
4) w/o fused features & 65.81 & 68.57 & 72.02 & 74.72 & 77.43 \\ 
\bottomrule
\end{tabular}
\caption{{\bf Contribution of different components of our model}. The values represent the average classification accuracies (\%) across the 11 datasets.}
\label{tab:ablations}
\end{table}
}

\section{Experiments}
\label{sec:experiments}
\subsection{Implementation Details}
We conducted experiments with the Causal CLIP Adapter (CCA) and its fine-tuned version (CCA-FT) on 11 widely used benchmark datasets covering various image classes, including ImageNet \cite{deng2009imagenet}, Caltech101 \cite{fei2004learning}, DTD \cite{cimpoi2014describing}, EuroSAT \cite{helber2019eurosat}, FGVC Aircraft \cite{maji2013fine}, Flowers102 \cite{nilsback2008automated}, Food101 \cite{bossard2014food}, Oxford Pets \cite{parkhi2012cats}, Stanford Cars \cite{krause20133d}, SUN397 \cite{xiao2010sun}, and UCF101 \cite{soomro2012ucf101}. We applied CLIP's preprocessing methods to prepare the training and validation/testing images and used ResNet50 as the CLIP backbone unless otherwise stated. 

During the construction of the disentangled cache model, we generated image features by averaging the CLIP features of 10 augmented views for each training image. The unmixing matrix $\mathbf{U}$ was extracted using FastICA, as implemented in \textsc{scikit-learn} \cite{sklearn_api}, on a randomly sampled subset of ImageNet \cite{deng2009imagenet}, consisting of 100,000 images (100 per class). It is calculated before training and used across all datasets. The resulting feature dimension was set to $M = 1024$, unless otherwise stated. For experiments using different CLIP backbones, the unmixing matrix $\mathbf{U}$ was calculated using the corresponding features from the same ImageNet subset, with the source dimension matching the CLIP feature dimension, i.e., $M = C$. 
For each test image, it was first disentangled using the pre-computed unmixing matrix $\mathbf{U}$ before calculating similarities $\mathbf{S}$ with the disentangled cache keys. The CLIP text classifier was initialized using the textual features of all class labels for each specific dataset that are embedded in prompt templates designed in \cite{zhang2022tip}.

We evaluated CCA in both the training-free setting and the fine-tuned setting with 1, 2, 4, 8, and 16 training samples. The CCA-FT was trained for 20 epochs for all datasets except EuroSAT, where it was trained for 100 epochs, following the settings in \cite{zhang2022tip}. During training, the CLIP encoders were kept frozen. The batch size was set to 256 for ImageNet and 128 for the other datasets. The learning rate for the cache adapter and CLIP text classifier was 0.001 and 0.0001, respectively, for all datasets. After training, for each dataset except ImageNet, the balancing factor $\alpha$, $\gamma$, $\eta$ and the smoothness factor $\beta$ were optimized on the corresponding validation set using grid search.

\subsection{Causal CLIP Adapter}
We evaluated the performance of CCA on 11 different downstream datasets and calculated the average classification accuracy. We compared the performance of CCA-FT with several state-of-the-art (SOTA) methods for CLIP-based few-shot learning, including APE\footnote{To ensure fair comparison, we used the prompt templates for Tip-Adapter \cite{zhang2022tip} instead of the prompt templates generated by GPT \cite{pratt2023does}.} \cite{zhu2023not}, TaskRes \cite{yu2023task}, cross-modal partial fine-tuning \cite{lin2023multimodality}, and LP++ \cite{huang2024lp++}. As shown in \cref{fig:11_datasets}, the training-free CCA has slightly lower average classification accuracy compared to the training-free Tip-Adapter. This could be attributed to the misalignment between disentangled image features within the same class, resulting from the linear transformation applied by the ICA unmixing matrix $\mathbf{U}$. However, after fine-tuning with the few-shot training samples, CCA-FT outperforms Tip-Adapter-F on all downstream datasets and surpasses all SOTA methods in terms of average accuracy across the 11 datasets. This suggests that disentangling the cache keys enhances optimization efficiency, while fine-tuning the CLIP text classifier and feature fusion improve inter-modal alignment, both contributing to the observed increase in classification accuracy.
\figAccs

\subsection{Generalization Performance}
To test our model's out-of-distribution robustness, we trained the CCA-FT using the ImageNet 16-shot training set as the source dataset and evaluated its performance on two variants of ImageNet, namely ImageNetV2 \cite{recht2019imagenet} and ImageNet-Sketch \cite{wang2019learning}, as the target datasets. We compared CCA-FT's performance with that of linear-probe CLIP \cite{radford2021learning}, CoOp \cite{zhou2022learning}, and Tip-Adapter. As shown in \cref{tab:ood}, 
disentanglement-based methods (i.e., either CCA or CCA-FT) achieve the best performance for both the source and the target datasets, suggesting that feature disentanglement enhances the model's robustness to distribution shifts.
\tabOOD

To evaluate our model's robustness to noise, we introduced Gaussian noise with parameters $\mu=0$ and $\sigma=0.2$ to the test images and evaluated the performance of CCA and CCA-FT on these noisy inputs. As presented in \cref{tab:noise}, CCA and CCA-FT demonstrate significantly greater resistance to noise compared to Tip-Adapter and Tip-Adapter-F, respectively. Additionally, to examine the model's resilience against adversarial attacks, we applied the Fast Gradient Sign Method (FGSM) \cite{goodfellow2014explaining} with $\epsilon=0.05$ to the test images and evaluated our models. The results, shown in \cref{tab:attack}, indicate that CCA and CCA-FT are substantially more robust to adversarial attacks than Tip-Adapter and Tip-Adapter-F, respectively.
\tabNoise
\tabAttack

\subsection{Computation Efficiency}
To evaluate our model's computational efficiency, we measure the 16-shot training time and performance on ImageNet using various methods including zero-shot CLIP, CoOp, and Tip-Adapter, on a single NVIDIA A100-SXM4-40GB GPU. Adapter-based methods such as CCA and Tip-Adapter pre-compute the CLIP textual features, whereas CoOp requires backpropagation through CLIP text encoder and recomputing the textual features after each gradient update. As a result, our method requires significantly less training time than CoOp, as shown in \cref{time}. Moreover, CCA achieves a greater performance improvement than CoOp without any training. CCA-FT outperforms Tip-Adapter-F with the same number of training epochs and a similar training time.
\tabTime

\subsection{Ablation studies}
\subsubsection{Effects of Different Components}
To evaluate the contribution of different components of our method, we analyzed the few-shot performance under four distinct ablation settings: (1) using the original CLIP features as cache keys without disentanglement by ICA, (2) freezing the cache adapter during training and only fine-tuning the CLIP text classifier, (3) freezing the CLIP text classifier during training and only fine-tuning the cache adapter, and (4) excluding the fused features generated by cross-attention. As shown in \cref{tab:ablations}, the first setting results in a significant drop in few-shot performance when the cache keys are not disentangled, suggesting that disentanglement plays a crucial role in mitigating overfitting and boosting performance. In the second setting, without fine-tuning the cache adapter, misalignment within the image modality persists, leading to degraded performance. For the third and fourth settings, freezing the CLIP text classifier or excluding the fused features results in a performance decrease compared to the full model, though to a lesser extent than in the first two settings. This may indicate that uni-directional and bi-directional cross-modal alignment can compensate for each other in our model. Overall, these findings suggest that different components of our method contribute to performance improvements to varying degrees.
\tabAblations

\subsubsection{Different Backbones}
\begin{figure}[ht]
    \centering
        \includegraphics[width=0.48\textwidth]{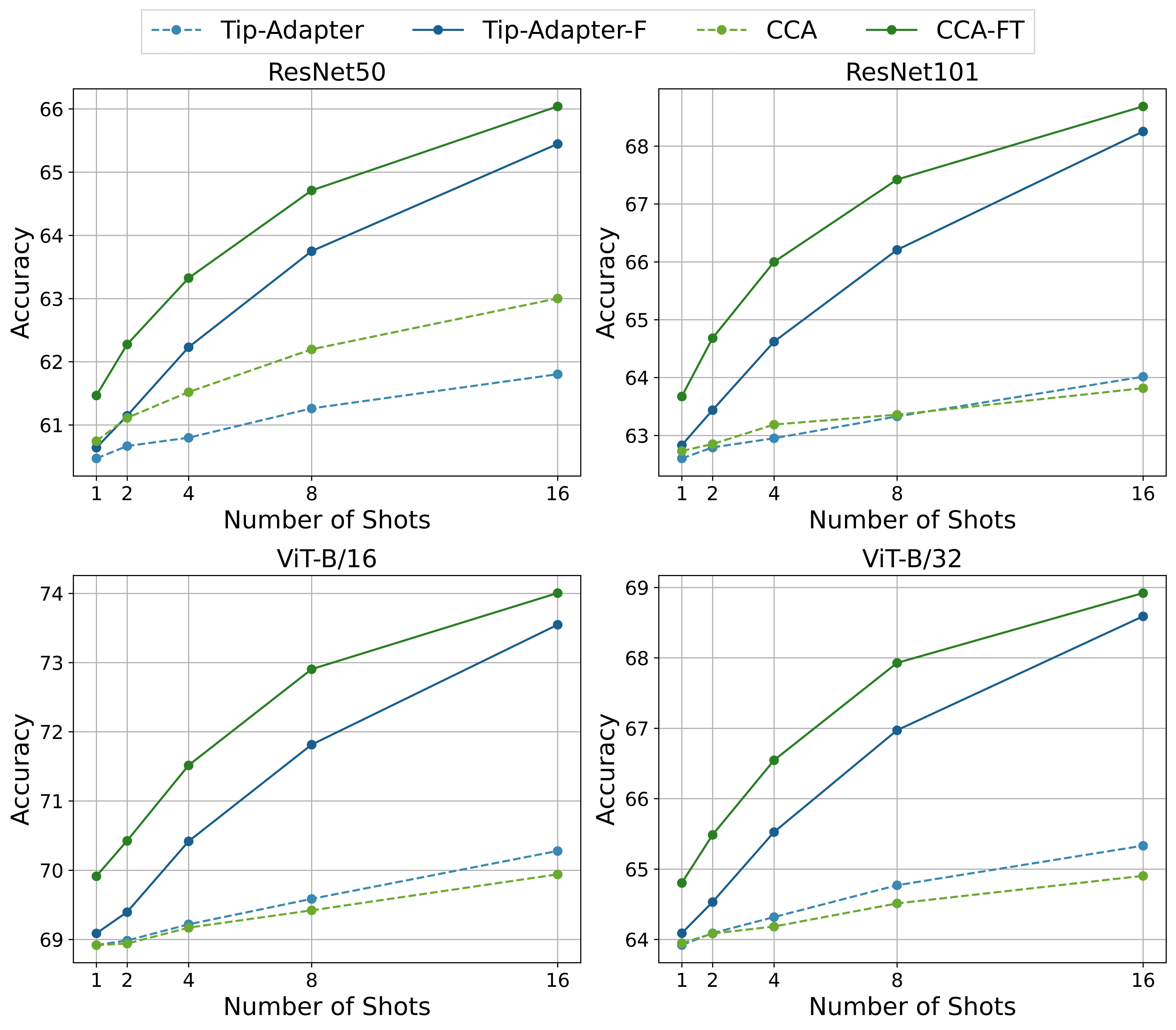}
    \caption{Comparison of classification accuracy between CCA(-FT) and Tip-Adapter(-F) for different CLIP encoders. The accuracy here represents 16-shot classification accuracy (\%) on ImageNet.}
    \label{fig:backbones}
\end{figure}

We evaluated our method on different types of CLIP backbones besides ResNet50, including ResNet101, ViT-B/16, and ViT-B/32, on ImageNet. As shown in \cref{fig:backbones}, the performance of the training-free CCA is lower than that of Tip-Adapter, possibly due to feature misalignment after ICA in the image modality. However, CCA-FT consistently outperforms Tip-Adapter-F across various CLIP backbones, demonstrating its robustness across different CLIP features.

\subsubsection{Hyper-parameter Sensitivities}
To assess the sensitivity of our model to the balancing factors $\alpha$, $\gamma$, $\eta$ and the smoothing factor $\beta$, we evaluated the performance of CCA and CCA-FT (trained on the ImageNet 16-shot training set) across varying hyper-parameter settings. As shown in \cref{fig:alpha} and \cref{fig:beta}, CCA and CCA-FT demonstrate a good balance between accuracy and robustness against different $\alpha$ and $\beta$ values, compared to Tip-Adapter and Tip-Adapter-F, respectively. As shown in \cref{fig:gamma_eta}, the model performance is relatively stable across a range of $\gamma$ and $\eta$ values, which might result from the mutually complementary cross-attention design.
\begin{figure}[ht]
     \centering
     \begin{subfigure}[b]{0.41\textwidth}
         \centering
         \includegraphics[width=\textwidth]{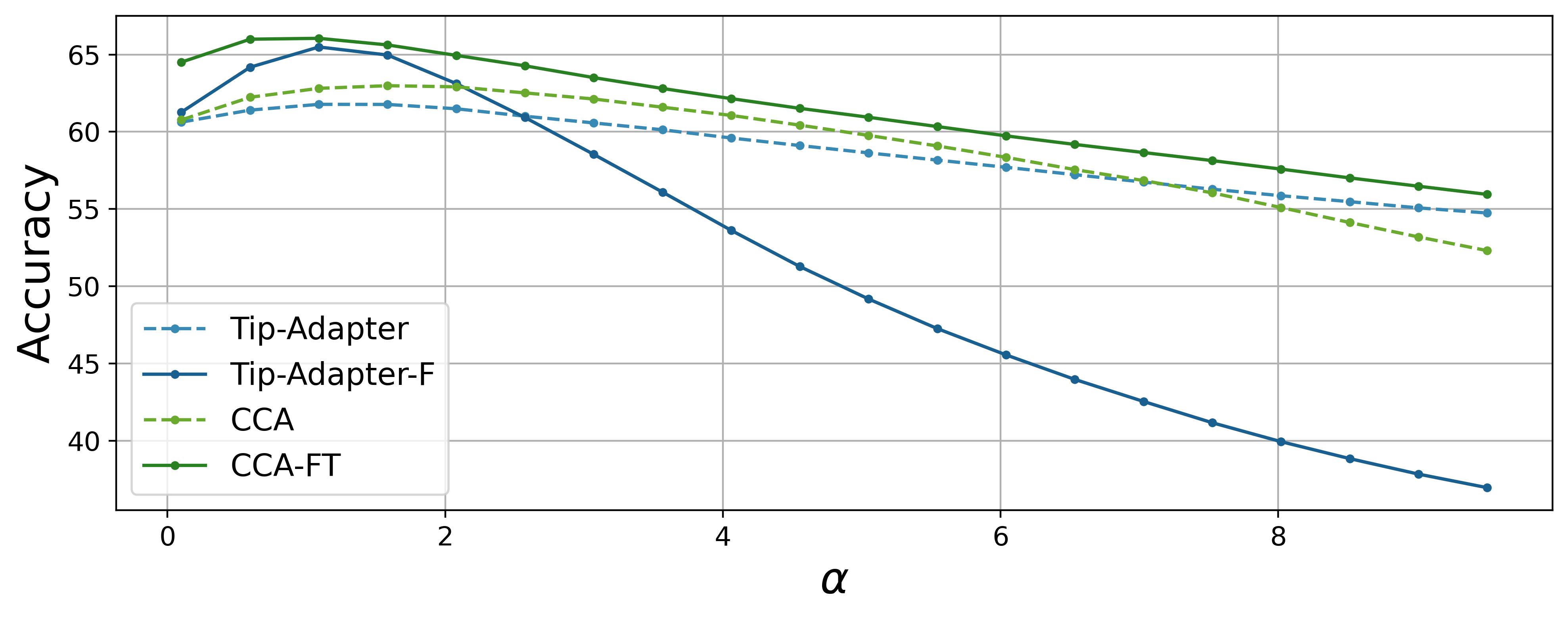}
         \caption{Sensitivity to $\alpha$.}
         \label{fig:alpha}
     \end{subfigure}
     \begin{subfigure}[b]{0.41\textwidth}
         \centering
         \includegraphics[width=\textwidth]{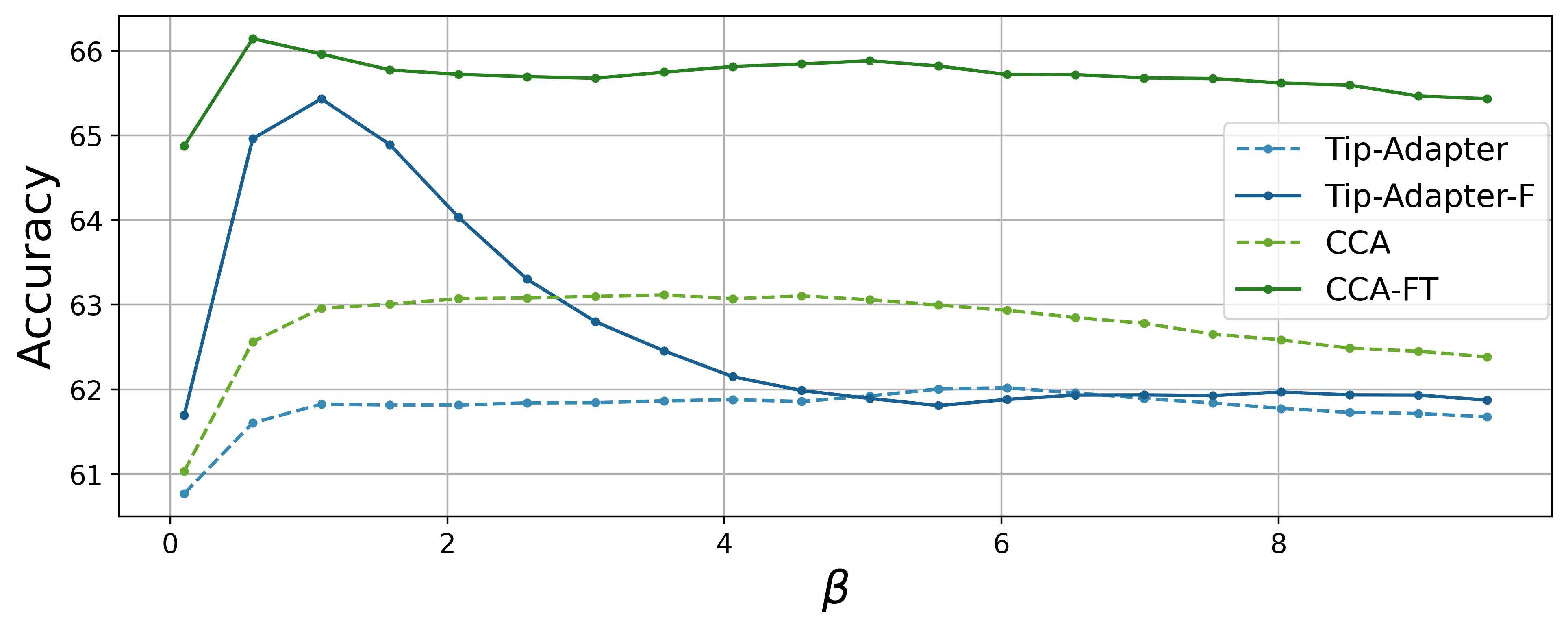}
         \caption{Sensitivity to $\beta$.}
         \label{fig:beta}
     \end{subfigure}
     \begin{subfigure}[b]{0.42\textwidth}
         \centering
         \includegraphics[width=\textwidth]{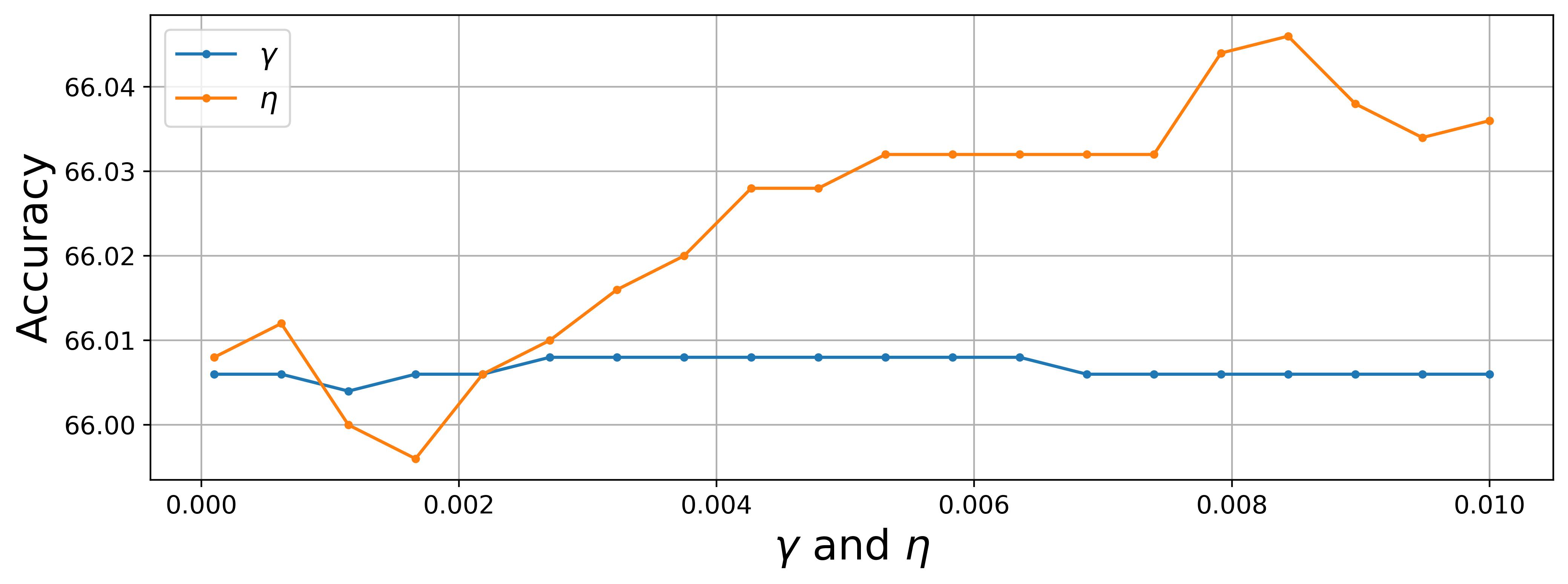}
         \caption{Sensitivity to $\gamma$ and $\eta$ for CCA-FT.}
         \label{fig:gamma_eta}
     \end{subfigure}
     \label{fig:hp}
     \caption{Sensitivities to the balancing factors $\alpha$, $\gamma$, $\eta$ and the smoothing factor $\beta$ for CCA(-FT) and Tip-Adapter(-F). The accuracy here represents 16-shot classification accuracy (\%) on ImageNet.}
\end{figure}


\section{Conclusion}
\label{sec:conclusion}
In this paper, we introduce the Causal CLIP Adapter (CCA), a novel method designed to enhance few-shot learning by leveraging causal disentanglement and cross-modal alignment. By applying ICA to CLIP features, our model effectively recovers independent latent variables, enabling better adaptation to distribution shifts.
To address misalignment introduced by ICA, we train a cache adapter and enhance cross-modal alignment both unidirectionally and bidirectionally by fine-tuning the CLIP text classifier and leveraging hybrid features generated via cross-attention.
Our method consistently outperforms SOTA approaches across 11 benchmark datasets in few-shot classification tasks and demonstrates superior robustness to distribution shifts. These results underscore the effectiveness of combining causal disentanglement with cross-modal alignment to tackle the challenges of few-shot learning.


\section*{Acknowledgement}
Z.Z. acknowledges the financial support from the Responsible AI Research Centre, Australian Institute for Machine Learning.

{
    \small
    \bibliographystyle{ieeenat_fullname}
    \bibliography{main}
}

\end{document}


\maketitle

\pagenumbering{gobble}
\setcounter{table}{5}
\setcounter{figure}{5}
\setcounter{section}{5}

\section{Comparison with Other Methods}
To allow a more comprehensive comparison with recent methods, we evaluated the performance of AMU-Tuning \cite{tang2024amu} and Transductive-CLIP \cite{martin2024transductive} alongside our own approach. As shown in Tab.~\ref{tab:additional}, our method consistently outperforms AMU-Tuning. Compared to Transductive-CLIP, our method achieves better performance in the 1-shot and 2-shot settings, but falls short in the 4-, 8-, and 16-shot settings.

It is important to note that Transductive-CLIP is highly sensitive to the number of classes in the query batch (denoted as $k_{\textit{eff}}$); its performance can drop significantly with even a slight increase in $k_{\textit{eff}}$. Moreover, as discussed in Related Work, AMU-Tuning incorporates an additional pre-trained model, MoCo v3 \cite{chen2021empirical}, alongside CLIP, rendering direct comparisons somewhat unfair. Similarly, Transductive-CLIP operates under a \textit{transductive setting}, where predictions are made jointly for a batch of samples, leveraging inter-sample dependencies. In contrast, our method uses a more general \textit{inductive setting}, performing inference independently for each sample. This difference in inference paradigm makes direct comparison inherently inequitable.

\begin{table}[h!]
\caption{Average few-shot classification accuracy (\%) across 11 datasets for different methods.}
\label{tab:additional}
\centering
\resizebox{0.48\textwidth}{!}{\begin{tabular}{l@{\hskip 0.11in}c@{\hskip 0.11in}c@{\hskip 0.11in}c@{\hskip 0.11in}c@{\hskip 0.11in}cc}
\toprule
Method & 1-shot & 2-shot & 4-shot & 8-shot & 16-shot \\ 
\midrule
AMU-Tuning & 64.50 & 66.95 & 70.57 & 72.87 & 74.71 \\ 
Transductive-CLIP ($k_{\textit{eff}}=5$) & 65.25 & 68.58 & 73.77 & 77.98 & 81.25 \\ 
Transductive-CLIP ($k_{\textit{eff}}=7$) & 61.60 & 64.08 & 68.99 & 73.58 & 76.97 \\ 
\midrule
CCA-FT (Ours) & 66.00 & 68.62 & 72.10 & 74.84 & 77.60 \\ 
\bottomrule
\end{tabular}}
\end{table}

\section{Ablation Study: Different ICA Dimensions}
\begin{figure}[ht]
    \centering
        \includegraphics[width=0.4\textwidth]{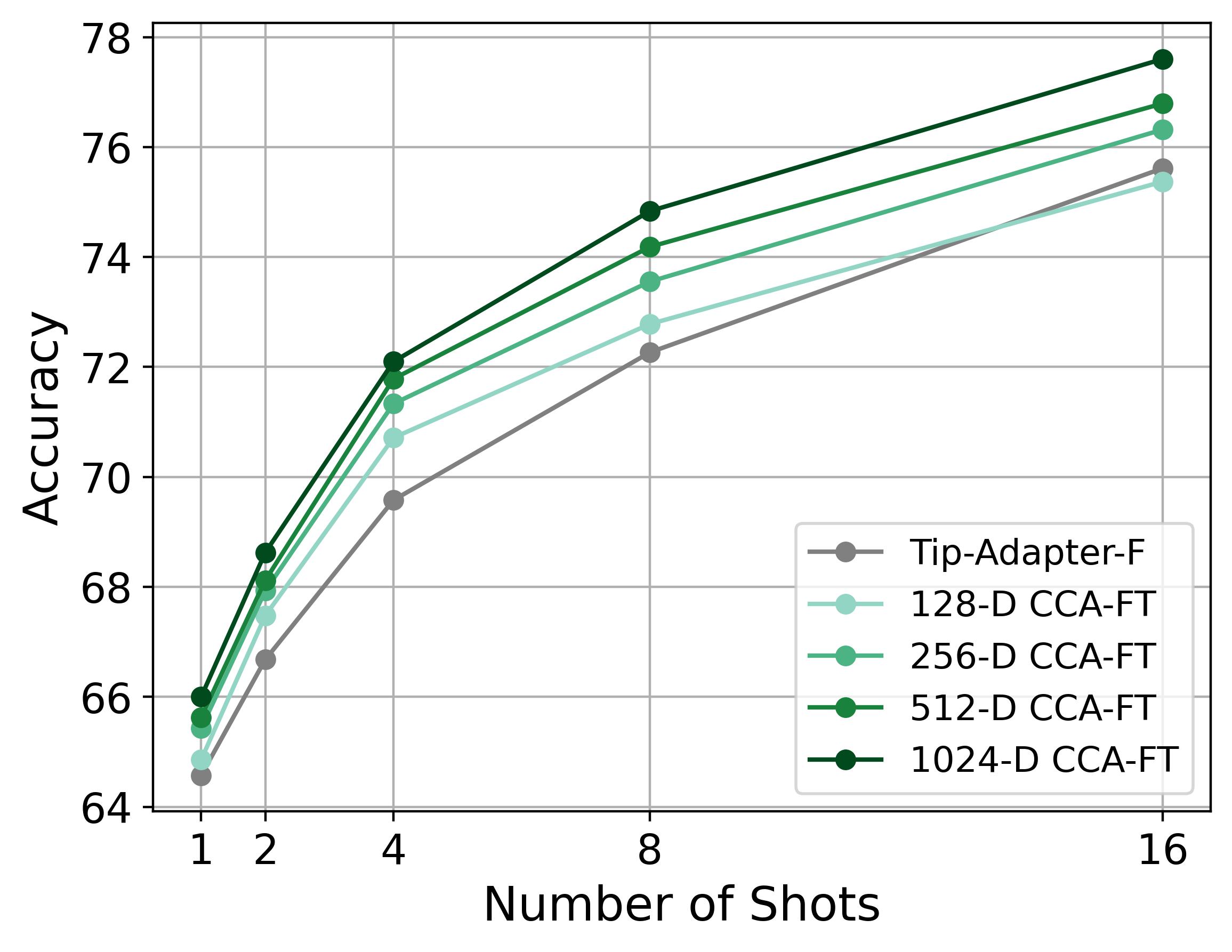}
    \caption{Average classification accuracy (\%) of CCA-FT with varying feature dimensions $M$ across 11 datasets, compared with Tip-Adapter-F.}
    \label{fig:ica_dim}
\end{figure}

To assess the robustness of CCA-FT with respect to varying ICA dimensions $M$, we perform few-shot experiments for $M \in \{128, 256, 512, 1024\}$ and compare the results with Tip-Adapter-F. As illustrated in \cref{fig:ica_dim}, CCA-FT consistently surpasses Tip-Adapter-F, even when the ICA dimension is as low as 128.

{
    \small
    \bibliographystyle{ieeenat_fullname}
    \bibliography{main}
}